\documentclass[runningheads]{llncs}
\usepackage{graphicx}
\usepackage{wrapfig}

\usepackage{tikz}
\usepackage{comment}
\usepackage{amsmath,amssymb} 
\usepackage{color}
\usepackage[colorlinks]{hyperref}
\usepackage{booktabs}
\usepackage{multirow}
\usepackage{multicol}
\usepackage{xcolor}
\usepackage{pifont}

\usepackage[accsupp]{axessibility}  

\begin{document}
\sloppy\hyphenpenalty=2000
\pagestyle{headings}
\mainmatter
\def\ECCVSubNumber{4292}  

\title{Towards Unbiased Label Distribution Learning for Facial Pose Estimation Using\\ Anisotropic Spherical Gaussian} 

\titlerunning{Unbiased Label Distribution Learning for Facial Pose Estimation}
%
\author{ Zhiwen Cao \inst{1}$^{\star}$
\and Dongfang Liu \inst{2}\thanks{Equal contributions.} 
\and Qifan Wang\inst{3}\thanks{The analysis and all work described in this paper was performed by the authors at Purdue and RIT. Qifan Wang served as an advisor to the project.}%
\and Yingjie Chen\inst{1}}
\authorrunning{Z. Cao et al.}
%
\institute{Purdue University \and
Rochester Institute of Technology \and
Meta AI\\
\email{\{cao270, victorchen\}@purdue.edu\\
dongfang.liu@rit.edu \\
wqfcr@fb.com}} 
\maketitle

\begin{abstract}
Facial pose estimation refers to the task of predicting face orientation from a single RGB image. It is an important research topic with a wide range of applications in computer vision. Label distribution learning (LDL) based methods have been recently proposed for facial pose estimation, which achieve promising results. However, there are two major issues in existing LDL methods. First, the expectations of label distributions are biased, leading to a \textit{biased pose estimation}. Second, \textit{fixed} distribution parameters are applied for all learning samples, severely limiting the model capability. 
In this paper, we propose an Anisotropic Spherical Gaussian (ASG)-based LDL approach for facial pose estimation. In particular, our approach adopts the spherical Gaussian distribution on a unit sphere which constantly generates \textit{unbiased expectation}. Meanwhile, we introduce a new loss function that allows the network to learn the distribution parameter for each learning sample \textit{flexibly}. Extensive experimental results show that our method sets new state-of-the-art records on AFLW2000 and BIWI datasets. 
\end{abstract}

\section{Introduction}
\begin{figure}
    \centering
    \includegraphics[width=0.45\textwidth]{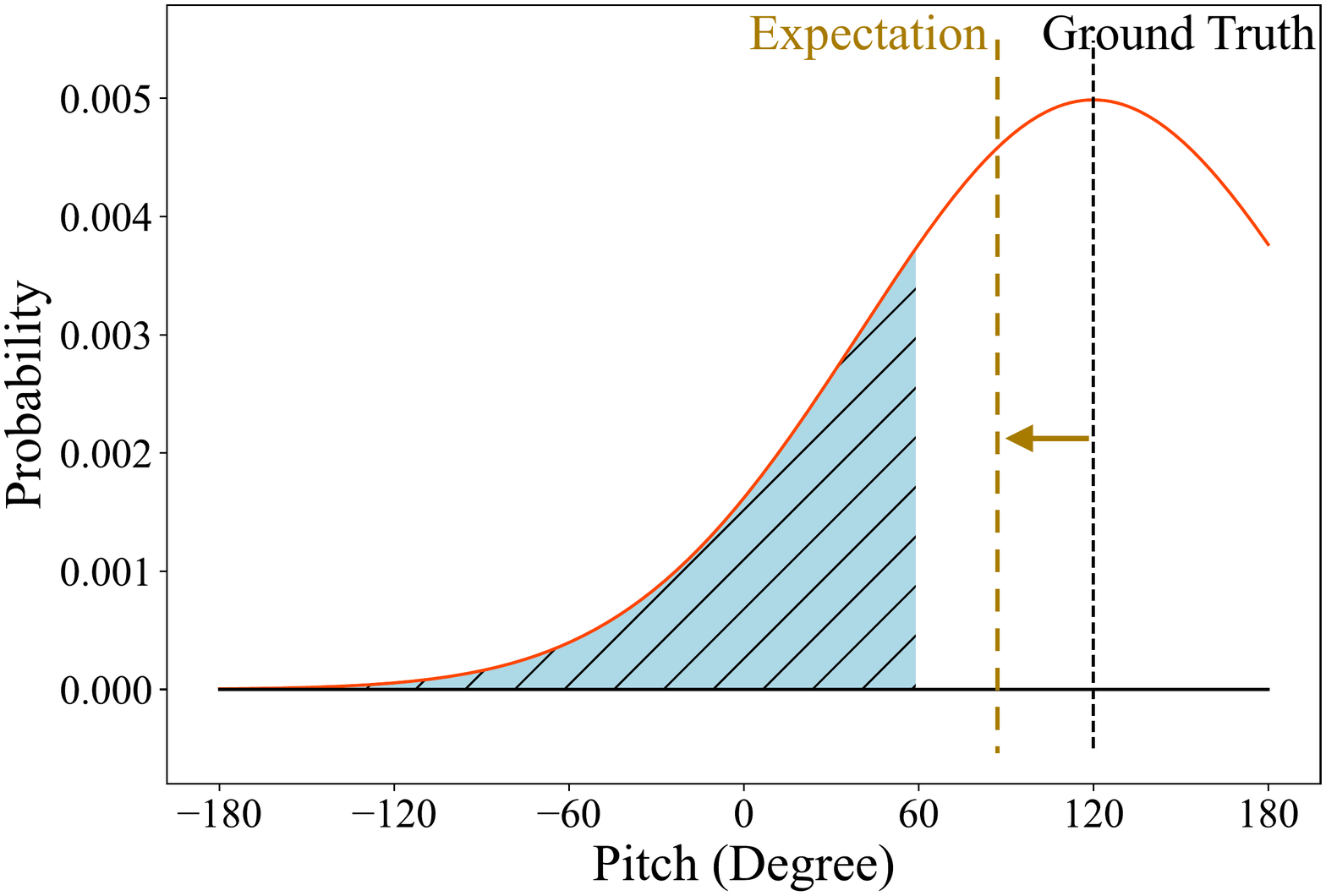}\includegraphics[width=0.55\textwidth, trim={0cm 0.3cm 0cm 0.2cm},clip]{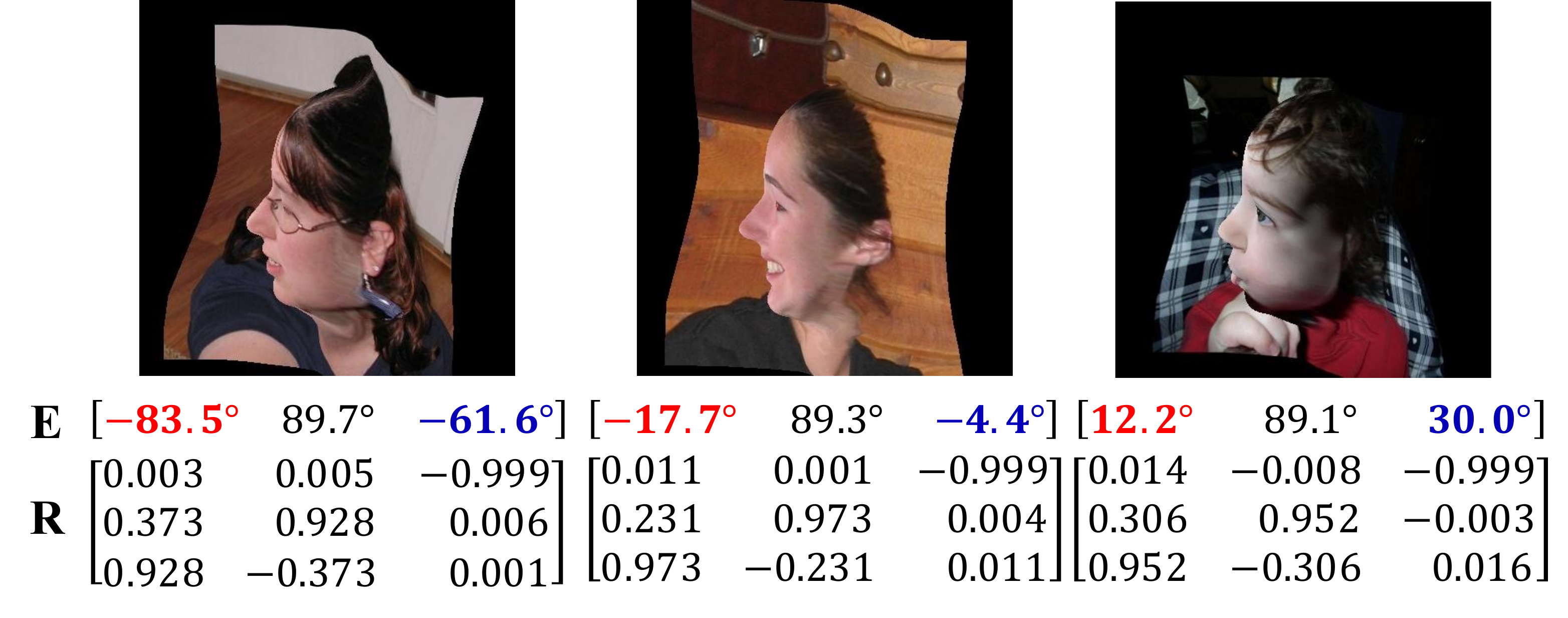}\\
    \text{(a)} \space \qquad \qquad \qquad \qquad \qquad \qquad \qquad \qquad \qquad \text{(b)}
    \caption{\textbf{(a)} \textbf{Example of Biased Expectation} for 1D Gaussian label distribution. The distribution from original pitch $=120^\circ$ in the range $(-180^\circ, +180^\circ]$ gives a biased expectation of the label. The condition becomes worse as the original angle gets closer to $180^\circ$ or $-180^\circ$. \textbf{(b)} \textbf{E} and \textbf{R} denote \textbf{\textit{Euler angles}} and \textbf{\textit{rotation matrix}}, respectively. Euler angles have inconsistent representations for profile faces. The \textcolor{red}{Red} and \textcolor{blue}{Blue} values show evident discrepancies when denoting similar profile poses. Applying LDL on Euler angles inevitably introduces heavily noisy supervision. In contrast to Euler angles, the corresponding elements in rotation matrices are close to each other. All samples are from the 300W-LP dataset~\cite{yang2016wider}.}
    \label{fig:euler_issue}
\end{figure}
The task of facial pose estimation is to estimate the orientation of the face from a single RGB image. It plays an important role in many real-world applications, including driver's monitoring system~\cite{geronimo2009survey,murphy2007head}, human-computer interaction~\cite{chen2019realistic,liu2019video} and face alignment~\cite{chang2017faceposenet,yang2015face}. With the recent advance of deep learning in computer vision~\cite{he2016deep,cheng2022physical,yan2022gl,liu2021densernet,cui2021tf,liu2021sg}, learning-based facial pose estimation has become a dominant approach, achieving promising results~\cite{albiero2020img2pose,Cao_2021_WACV,ruiz2018fine,valle2020multi}. However, as a general problem in deep learning,  data shortage also limits the concurrent methods for facial pose estimation to achieve superior performance. How to effectively estimate the facial pose with limited data remains a challenge, which is the focus of this work.

Recently, label distribution learning (LDL) has been introduced to address the issue of insufficient training data. These LDL methods aim at reconstructing new labels of the distribution around the original ones for training, which promote the learning of facial images not only from their own labels but also the adjacent ones. LDL has shown its effectiveness in tasks such as facial age estimation~\cite{6475129}, facial attractiveness estimation~\cite{fan2017label} and crowd counting~\cite{zhang2015crowd}. However, the exploration of LDL application to facial pose estimation is insufficient. 
\begin{figure}[t]
    \centering
    \includegraphics[clip,trim=2.5cm 5.5cm 3cm 6cm, width=0.8\textwidth]{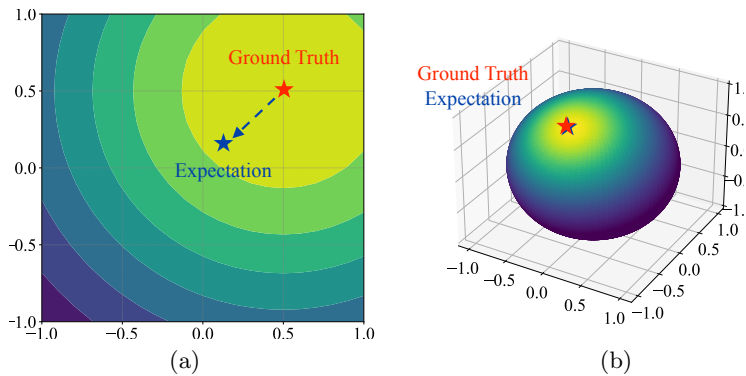}\\
    \qquad \text{(a)} \qquad \qquad \qquad \qquad \qquad \qquad \qquad  \qquad \text{(b)}
    \caption{\textbf{(a)} \textbf{2D distribution} \cite{Cao_2021_WACV} generated by the first two elements of a column vector of rotation matrices (we omit the third element for visualization). The limit of range $[-1, 1]$ in two directions results in biased expectation. \textbf{(b)} \textbf{Spherical Gaussian Distribution} (ours) guarantees \textbf{\textit{unbiased expectation}}.}
    \label{fig:2d_sphere_compare}
\end{figure}

To date, LDL in the task of facial pose estimation are mainly applied on Euler angles which are known as pitch, yaw and roll. A seminal work from \cite{geng2014head} proposed to use a 2D Gaussian Distribution to describe the probability distribution between pitch and yaw in the range of $(-90^\circ, +90^\circ)$. Liu $et$ $al.$~\cite{liu2019facial} followed the track and converted each Euler angle label to a 1D Gaussian distribution. They also expand the task to the one of wild range. Therefore, each face image corresponds to three 1D Gaussian distributions ($i.e.,$ $\rho_{pitch}, \rho_{yaw}$ and $\rho_{roll}$). For instance, an original label of the pitch angle $(120^{\circ})$ can be used to generate a Gaussian distribution in $(-180^\circ, 180^\circ]$ (see Fig.~\ref{fig:euler_issue}a). Through predicting the probability of each integer degree in set $\mathcal{S} = \{-179^\circ, -178^\circ, \cdots, 179^\circ, 180^\circ\}$ and compute the cross entropy loss, the task can be considered as the combination of both regression and classification.

Albeit being simple and effective, applying LDL on Euler angles has several obstacles: \textbf{\large\ding{172}} Euler angle is not a continuous rotation representation and LDL deteriorates the issue by contributing the learning of adjacent angles. The discontinuity is embodied in the Euler angle labels of profile faces (see Fig.~\ref{fig:euler_issue}b). Since similar profile images have very different Euler angle labels, converting the angles to distributions cannot help the learning of adjacent labels; \textbf{\large\ding{173}} Gaussian distribution labeling on Euler angles leads to biased expectations. Since the angle is limited to a certain range $(-180^{\circ}, 180^{\circ}]$, probabilities assigned in the the shadow area make the expectation of labels incorrectly shift to left (see Fig.~\ref{fig:euler_issue}a); and \textbf{\large\ding{174}} concurrent LDL methods utilize the variance of Gaussian distribution as a hyper-parameter which is fixed during training. This is computationally inefficient because they need to perform an exhaustive search to cherry pick the best parameter setting. Besides, using the same distribution for all the poses is not aligned with the real situation. Since faces at different poses have diverse contributions to adjacent faces, the network should learn the distribution parameters adaptively.

The first issue was studied in \cite{Cao_2021_WACV}, which identifies the discontinuity issue of Euler angle and proposed a vector-based representation to train the network. In other words, they let the neural network learn the rotation matrices from facial images. Rotation matrices can form a continuous special orthogonal group $SO(3)$ and can circumvent the problem of discontinuity. However, they still failed to recognize the issue of biased expectation. Since every element of the rotation matrix stays in the range of $[-1, 1]$, they convert each element to a Gaussian distribution in range $[-1, 1]$ and let the network learn the distribution in an element-wise manner. Consequently, the issue of biased expectation is inevitably similar to Euler angles. To our best knowledge, the second and third issues remain largely under-explored in existing literature.

In light of the foregoing discussions, we are motivated to present our Anisotropic Spherical Gaussian (ASG)-based label distribution learning method for facial pose estimation. Specifically, we treat each column vector of the rotation matrix as an entity and map them to a spherical Gaussian distribution respectively. Due to the symmetric distribution of ASG, our approach guarantees for an unbiased expectation of label distributions. The difference between ASG and the method in \cite{Cao_2021_WACV} is demonstrated in Fig.~\ref{fig:2d_sphere_compare}.
Armed with the spherical Gaussian distribution, we further design a new loss function for the network to learn the distribution parameters adaptively during the training stage. This enables every facial image to adjust contribution to adjacent poses based on its pose. 
Ablation studies show that it can transcend the cherry-picked parameter by at least $4.0\%$ when trained on 3000W-LP and tested on AFLW2000 dataset.\\
\indent Our method enjoys a few attractive qualities: \textbf{\large\ding{182}} it ensures the network learns the distribution with \textbf{\textit{unbiased expectation}}. Since most existing methods have biased expectations (unless the original ground truth is exactly in the middle), we observe significant performance gain from our method; \textbf{\large\ding{183}} the capacity of learnable ASG distribution parameters allows the network to adjust the parameter for each pose, enabling a \textbf{\textit{fine-grained}} prediction;
\textbf{\large\ding{184}} all the performance achievement comes from optimization on \textbf{\textit{representation of rotation}} without increasing the size of neural networks. Our approach achieves state-of-the-art performance with a very light-weighted backbone network, $i.e.,$ ResNet18~\cite{he2016deep}. Specifically, we decrease the Mean Absolute Error (MAE) by $0.27^\circ$ ($6.9\%\downarrow$) compared to \cite{albiero2020img2pose} and $0.19^\circ$ ($5.0\%\downarrow$) compared to \cite{valle2020multi} when tested on AFLW2000 dataset \cite{zhu2016face}; and 
\textbf{\large\ding{185}}, our method is the \textbf{\textit{first attempt}} that adopts directional statistics in the task of pose estimation. We believe it can help invoke more thoughts for further exploration in the community.
Our contributions are summarized below:
\begin{itemize}
    \item We propose a novel ASG-LDL method which encodes each column vector of the rotation matrix as an anisotropic spherical gaussian on a unit sphere. Our method addresses the issue of biased expectation that is under-explored in previous works.
    \item We propose a novel training paradigm that allows the network to learn the distribution parameters adaptively. 
    The flexibility allows the network to learn individual distribution parameters for each pose.
    \item We conduct extensive experiments on two benchmarks. Experimental results show the effectiveness of our method. With a light-weight ResNet-18 as the backbone, our method achieves state-of-the-art results and outperforms many strong baselines with a heavier backbone ($i.e.,$ ResNet-50).
\end{itemize}

\section{Related Work}

This section summarizes the recent progresses in the related fields regarding facial pose estimation, label distribution learning and spherical Gaussian distribution.\\
\noindent\textbf{Facial pose estimation.} Recently, landmark-free learning based methods have become popular. By training an end-to-end deep neural network, it can estimate the face poses using global information and can be more robust to the environment variations. \cite{ruiz2018fine} puts forward a CNN with a multi-loss function that performs binned classification to regress three Euler angles. \cite{yang2019fsa} proposes a fine-grained structure by learning global spatial feature importance that improves the results. \cite{hsu2018quatnet} formulates face pose estimation using quaternion-annotated labels to avoid the ambiguity problem in Euler angle representation. \cite{albiero2020img2pose} proposes a Faster RCNN based network to regress 6DoF pose of faces by performing pose estimation and face alignment simultaneously. \cite{valle2020multi} puts forward a multi-modal network that can perform three tasks of head pose estimation, landmark-based face alignment and localization of face simultaneously. By the combination of three tasks they achieve state-of-the-art results. All of the above methods perform the training process through direct regression. Differently, we approach the problem as a label distribution learning task.\\
\noindent\textbf{Label distribution learning.} Label distribution learning~\cite{liu2021unveiling} is a learning paradigm that is first proposed for facial age estimation~\cite{6475129,7406391}. ~\cite{6475129} finds that faces at close ages look similar. Therefore, they map each face image to a label distribution which covers a certain number of ages. Through this way one face image can contribute to not only the learning of its chronological age, but also the learning of its adjacent ages. LDL also shows it effectiveness in similar tasks such as facial attractiveness estimation~\cite{fan2017label}, crowd counting~\cite{zhang2015crowd} and movie rating prediction~\cite{geng2015pre} $etc$. \cite{diaz2019soft} applies a similar approach on ordinal regression such as image ranking and monocular depth estimation. \cite{Cao_2021_WACV} shows that the evaluation metric, mean absolute error of Euler angles (MAE), cannot reflect the actual performance especially for profile faces. Instead, they propose to use mean absolute error of vectors (MAEV) as a new metric. However, all the methods give biased expectation from the distribution, which severely limits the performance of neural network.\\
\noindent\textbf{Spherical Gaussian distribution.} Spherical Gaussian (SG) distribution, also known as von Mises-Fisher distribution~\cite{fisher1953dispersion}, is commonly used to simulate the properties of illumination and reflection in computer graphics. \cite{hara2005multiple} uses SG to estimate multiple light sources and reflectance properties. \cite{wang2009all} approximates the normal distribution function (NDF) by a mixture of SGs. \cite{de2011real} uses SG for the approximation of Bidirectional Transmittance Distribution Function (BTDF) for real-time estimation of environment lighting. However, SG only describes an isotropic distribution. \cite{xu2013anisotropic} further  proposes the ASG distribution which can describe an anisotropic distribution for rendering applications. Inspired by the above work, we successfully extend ASG to the field of head pose estimation.\\
\begin{figure}
\begin{center}
\includegraphics[width=0.95\linewidth]{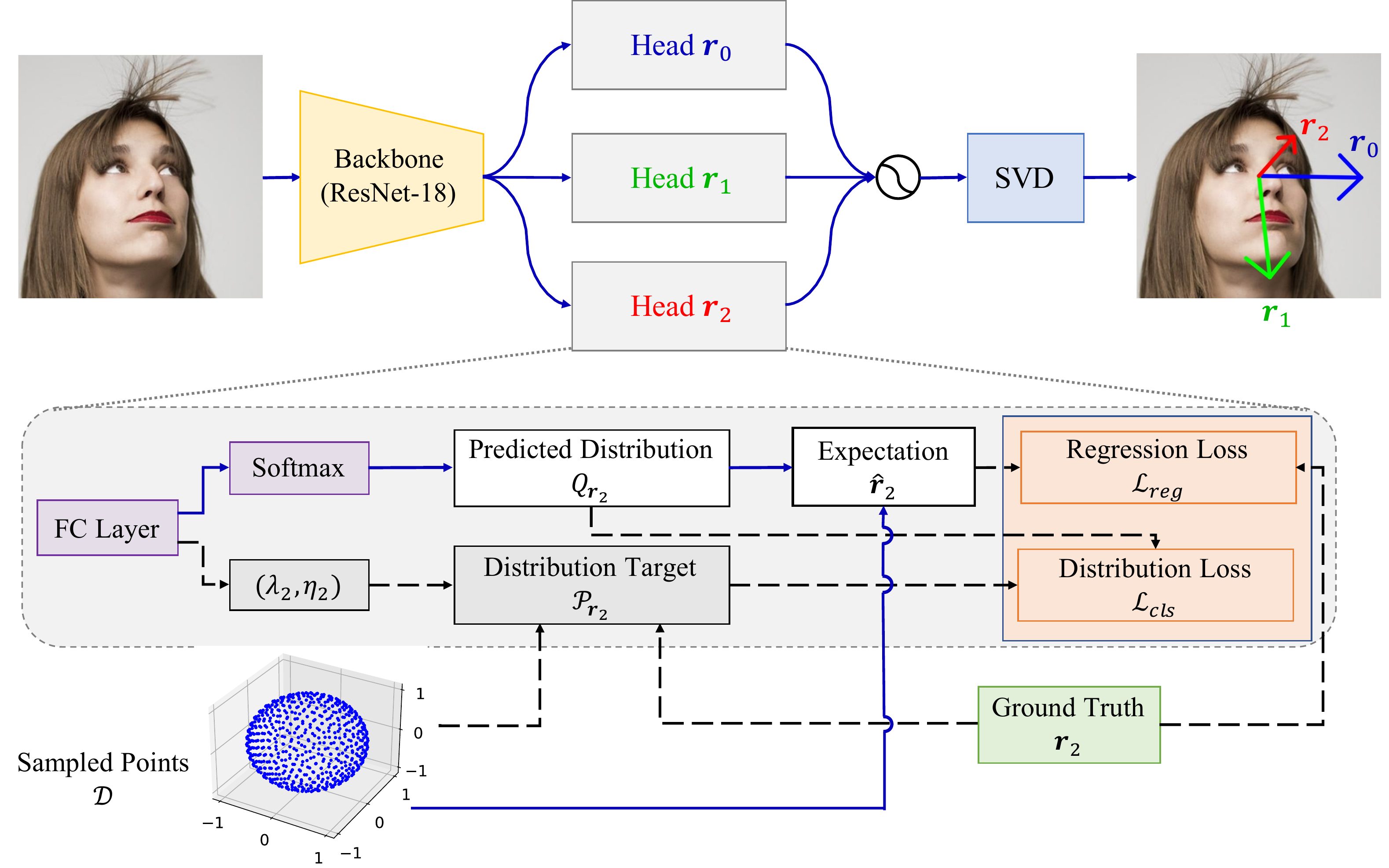}
\end{center}
   \caption{\textbf{The overall framework.} The dashed lines are only used in the training stage. For simplicity, we only visualize one head $\boldsymbol{r}_2$ but each head has the same working pipeline.}
\label{fig:NN}
\end{figure}
\noindent \textbf{6D object pose estimation.} 6D object pose estimation includes estimation of 3D location and 3D orientation. The latter task resembles our head pose estimation. The approaches for 6D object pose estimation can be generally classified into two categories. The first type such as \cite{peng2019pvnet,song2020hybridpose,zakharov2019dpod} first capture instance information and keypoints from images to determine locations of objects, then build the correspondence between the 2D and 3D keypoints. After that, they obtain 6D pose estimation by solving the PnP problem \cite{lepetit2009epnp}. The other category of methods such as \cite{gao2018occlusion,mahendran20173d,xiang2017posecnn} use neural networks to estimate orientation of objects directly. To our knowledge, the use of spherical Gaussian is a new attempt in the task of pose estimation.

\section{Proposed Method}
\subsection{Overview}
\begin{figure}
    \centering
    \includegraphics[clip,trim=4cm 7.5cm 0cm 8cm, width=0.95\textwidth]{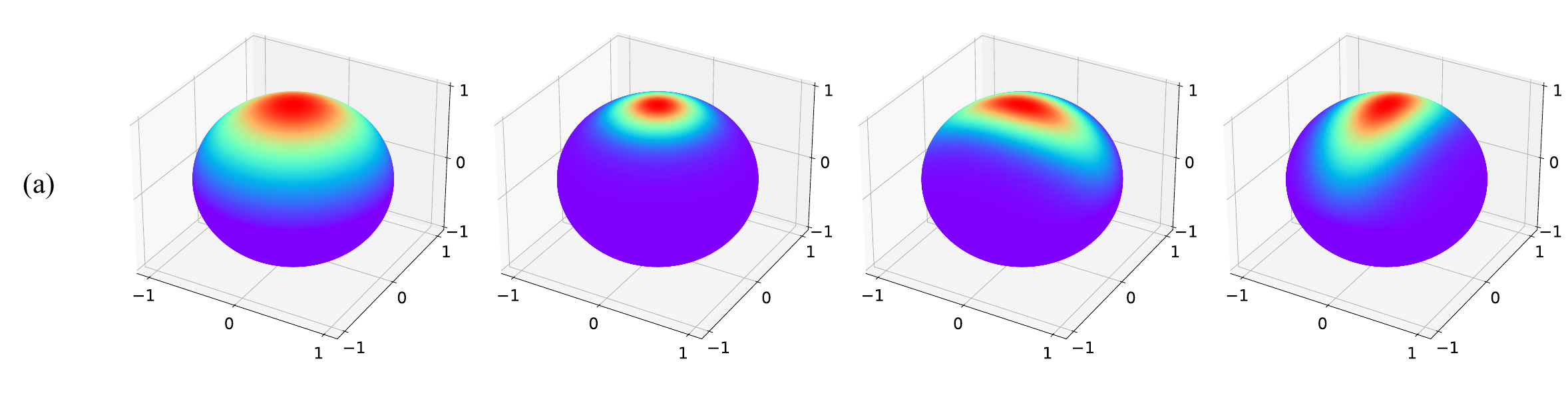}\\
    \includegraphics[clip,trim=3.65cm 0cm 0cm 0cm, width=0.95\textwidth]{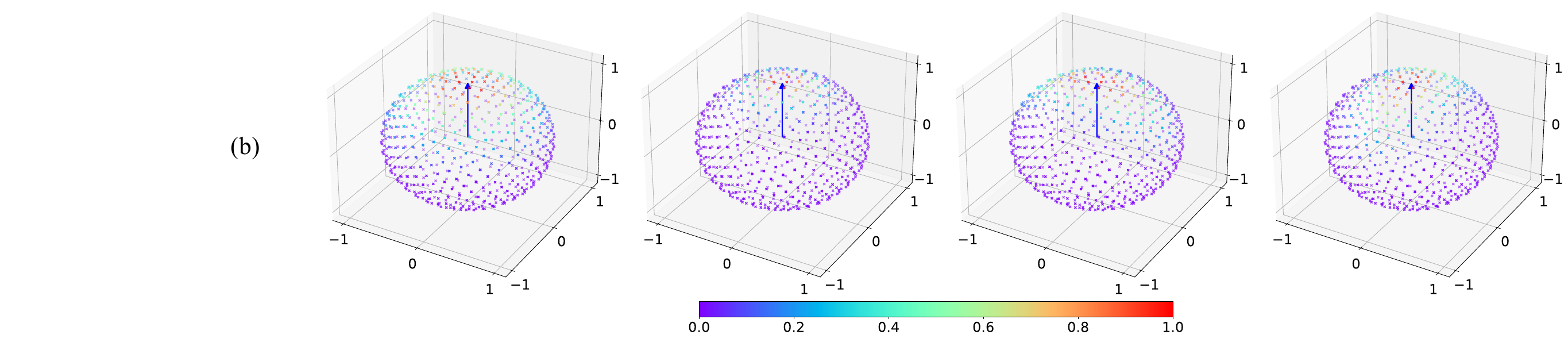}\\
     \qquad $\lambda=1, \eta =1$ \qquad \space $\lambda=5, \eta =5$ \qquad \space $\lambda=1, \eta =5$ \qquad \space $\lambda=5, \eta =1$
\caption{\textbf{Visualization of ASG distributions} of different $\lambda$ and $\eta$ when $\boldsymbol{r} = [0, 0, 1]^T$. a) ASG distribution on a unit sphere; b) visualization of sampled points with probabilities when $M=600$.}
\label{fig:sphere}
\end{figure}
Our overall framework is illustrated in Fig.~\ref{fig:NN}. The network learns the pose information from a cropped human facial image. To demonstrate the advantage of our approach, we choose light-weighted ResNet-18 as our backbone network. We append three heads to the ResNet-18 backbone as each head corresponds to one pose vector. They work collectively to perform the facial pose estimation. During training, the backbone first extracts the features from the input image and then feeds them to each of the heads, which is supervised by the classification and regression loss respectively. During inference, the three heads work collaboratively to predict the rotation matrix through singular value decomposition (SVD). We elaborate our method in the following sections.

\subsection{Motivation}
\indent When we use a rotation matrix $\boldsymbol{R}_{3\times 3} = \left[ \boldsymbol{r}_0, \boldsymbol{r}_1, \boldsymbol{r}_2 \right]$ to describe facial poses, the three column vectors $\boldsymbol{r}_0$, $\boldsymbol{r}_1$, $\boldsymbol{r}_2$ are equivalent to three pose vectors in Fig.~\ref{fig:NN}, $i.e.$ left (blue), down (green) and front (red) vectors respectively \cite{Cao_2021_WACV}. Therefore, for a ground truth pose vector $\boldsymbol{r}_i, i = \{0, 1, 2\}$, any direction $\boldsymbol v$ surrounding it can also be regarded as an alternative legitimate label. The smaller the angle difference is, the more likely that the vector $\boldsymbol v$ is a valid label. Therefore, all the probabilities of a direction $\boldsymbol{v}$, that can be considered as a legitimate label, constitute a probability distribution on a unit sphere. Intuitively, the probability distribution can be represented using an isotropic spherical Gaussian (SG) model, since the probability is only related to the angle between $\boldsymbol{v}$ and $\boldsymbol{r}_i$. However, human faces change at different rates when rotating along different axes. For example, rolling a face with $45^\circ$ does not change the observed area of the face, while nodding or raising the face for $45^\circ$ makes large portion of facial area self-occluded. Based on this observation, we propose to use ASG distribution which is able to capture the anisotropic features along different axes.

\subsection{Label Distribution Construction}
All three pose vectors constitute an orthogonal coordinate system. For each ground truth pose vector $\boldsymbol{r}_i$, we can calculate the portion $G^{i}$ that a direction $\boldsymbol{v}$ accounts for a full class description of the sample:

\begin{align}
    \label{eq:asg_pdf}
    G^{i}(\boldsymbol{v}; \boldsymbol{R}, &[\lambda, \eta])=c \cdot \mathrm{S}(\boldsymbol{v} ; \boldsymbol{r}_i) \cdot e^{-\lambda(\boldsymbol{v} \cdot \boldsymbol{r}_j)^{2}-\eta(\boldsymbol{v} \cdot \boldsymbol{r}_k)^{2}} \notag\\
    \text{where } &i = \{0, 1, 2\}\notag\\
    & j = (i+1)\bmod 3\notag\\
    & k = (i+2)\bmod 3. 
\end{align}

Here, $\boldsymbol{R} = [\boldsymbol{r_0}, \boldsymbol{r}_1, \boldsymbol{r}_2]$. $\lambda$ and $\eta$ are the parameters that control the decreasing speed of possibility along $\boldsymbol{r}_j$ and $\boldsymbol{r}_k$. Fig.~\ref{fig:sphere}a illustrates the spherical Gaussian distribution of different $\lambda$ and $\eta$. $c$ is the normalization term that ensures the sum of probability distribution to be 1. $S(\boldsymbol{v};\boldsymbol{r}_i)=\max(\boldsymbol{v}\cdot \boldsymbol{r}_i, 0)$ is the smooth term. Since the exponential part $B(\boldsymbol v)=e^{-\lambda(\boldsymbol{v} \cdot \boldsymbol{r}_j)^{2}-\mu(\boldsymbol{v} \cdot \boldsymbol{r}_k)^{2}}$, also known as Bingham distribution~\cite{kent1982fisher}, is antipodally symmetric and has two peaks at $\boldsymbol v = \pm \boldsymbol {r}_i$. We keep only the peak of $\boldsymbol v = \boldsymbol{r}_i$ with the smooth term $S(\boldsymbol v; \boldsymbol{r}_i)$.\\
\indent To convert a vector to a distribution, we first adopt spherical Fibonacci lattice algorithm \cite{gonzalez2010measurement} to sample $M$ near-equidistant points from an unit sphere, denoted by $\mathcal{D} = \{\boldsymbol d_1, \boldsymbol d_2, \cdots, \boldsymbol d_M\}$ where $\boldsymbol{d}_i \in \mathbb{R}^3$ (see Fig.~\ref{fig:NN}). Note that we only perform the sampling once, thus all pose vectors share a same set of sampled points. During the training stage, for any ground truth vector label $\boldsymbol{r}_i$, the network first predicts parameters $\lambda$ and $\eta$ and then use them to calculate the probabilities for all the sampled points $\mathcal{P}_{\boldsymbol{r}_i} = \{p_{1}^{i}, p_{2}^{i}, \cdots, p_{M}^{i}\}$. The probability of point $k$ can be obtained by the normalization:

\begin{align}
    \label{eq: normalized_asg}
    p_{k}^{i} = \frac{\exp\{\text{G}^{i}(\boldsymbol{v}_{k}; \boldsymbol{R}, [\lambda, \eta]\}}{\sum_{j=1}^{M} \exp\{\text{G}^{i}(\boldsymbol{v}_{j}; \boldsymbol{R}, [\lambda, \eta]\}}.
\end{align}

The process of label distribution generation is applied on all three column vectors $\boldsymbol{r}_{0}, \boldsymbol{r}_{1}$ and $\boldsymbol{r}_{2}$. Therefore, we can obtain three sets of probability distribution $\mathcal{P}_{\boldsymbol{r}_{0}}, \mathcal{P}_{\boldsymbol{r}_{1}}$ and $\mathcal{P}_{\boldsymbol{r}_{2}}$ with the same size of $M$. The probability distribution on sampled points are visualized in Fig.~\ref{fig:sphere}b.

\subsection{Working Pipeline}
\noindent\textbf{Training.} In the training stage, the backbone-encoded features are first fed into three heads separately (See Fig.~\ref{fig:NN}). Each head has one fully connected (FC) layer, which outputs a vector with size of $M+2$. 
The first $M$ elements denote the ASG probabilities of sampled points for the corresponding pose vector, which is normalized by a softmax layer to generate
the probability distribution
$\mathcal{Q}_{\boldsymbol{r}_i} = \{q_1^i, q_2^i, \cdots, q_M^i \}$. Therefore the expectation of the distribution is given by:

\begin{align}
    \hat{\boldsymbol r}_i = \mathbb{E}_{\mathcal Q_{\boldsymbol{r}_{i}}}[\mathcal D] = \sum_{k=1}^{M} q_{k}^{i} \boldsymbol d_{k}. 
    \label{eq:expectation}
\end{align}

The last two elements of the output vector from the FC layer correspond to the parameters $(\lambda_{i}, \eta_{i})$. In conjunction with the sampled point set $\mathcal{D}$ and the ground truth vector $\boldsymbol{r}_{i}$, the network is able to generate the distribution target $\mathcal{P}_{\boldsymbol{r}_{i}}$ using Eq.~\ref{eq:asg_pdf} and~\ref{eq: normalized_asg}.\\
\noindent \textbf{Loss function.} To supervise our method, our training loss consists of two terms: classification loss $\mathcal{L}_{cls}$ and regression loss $\mathcal{L}_{reg}$. The overall loss $\mathcal{L}$ is given by:
\begin{align}
    \mathcal{L} = \mathcal{L}_{cls} + \alpha L_{reg}.
\end{align}

More concretely, we adopt mean square error (MSE) loss function for regression $\mathcal{L}_{reg} = \text{MSE }(\boldsymbol{r}_i, \hat{\boldsymbol{r}}_i)$ and Kullback-Liebler (KL) divergence for classification $\mathcal{L}_{cls} = \text{D}_{\text{KL}}(\mathcal{P}_{\boldsymbol{r}_{i}} || \mathcal{Q}_{\boldsymbol{r}_{i}})$. The value of the trade-off parameter $\alpha$ is in the range of $[0, 1]$. We find its optimal value in our experiments.\\
\noindent\textbf{Inference.} In the inference stage, we first concatenate the three pose vectors $\hat{\boldsymbol{r}}_0, \hat{\boldsymbol{r}}_1, \hat{\boldsymbol{r}}_2$ generated by the three heads from the learned network to obtain matrix $\hat{\boldsymbol{R}} = \left[ \hat{\boldsymbol{r}}_0, \hat{\boldsymbol{r}}_1, \hat{\boldsymbol{r}}_2 \right]$. We then obtain its closest rotation matrix through singular value decomposition (SVD). Given a matrix $\hat{\boldsymbol R} = \boldsymbol U \boldsymbol \Sigma \boldsymbol V^T $, its closest rotation matrix is obtained by $\boldsymbol R = \boldsymbol U \boldsymbol V^T$.

\begin{figure}[t]
\begin{center}
    \includegraphics[clip,trim=1.05cm 0cm 0cm 0cm,width=3.5cm]{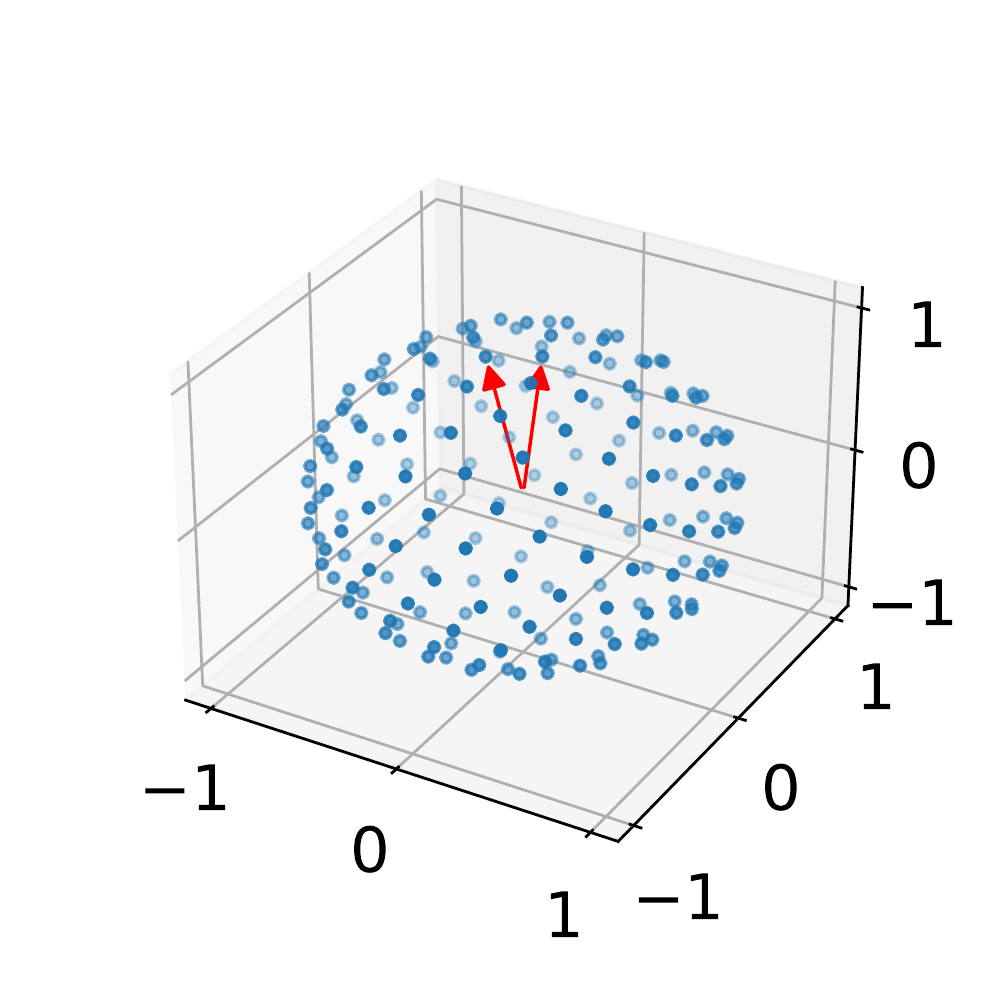} \qquad
    \includegraphics[width=5cm,height=4.1cm]{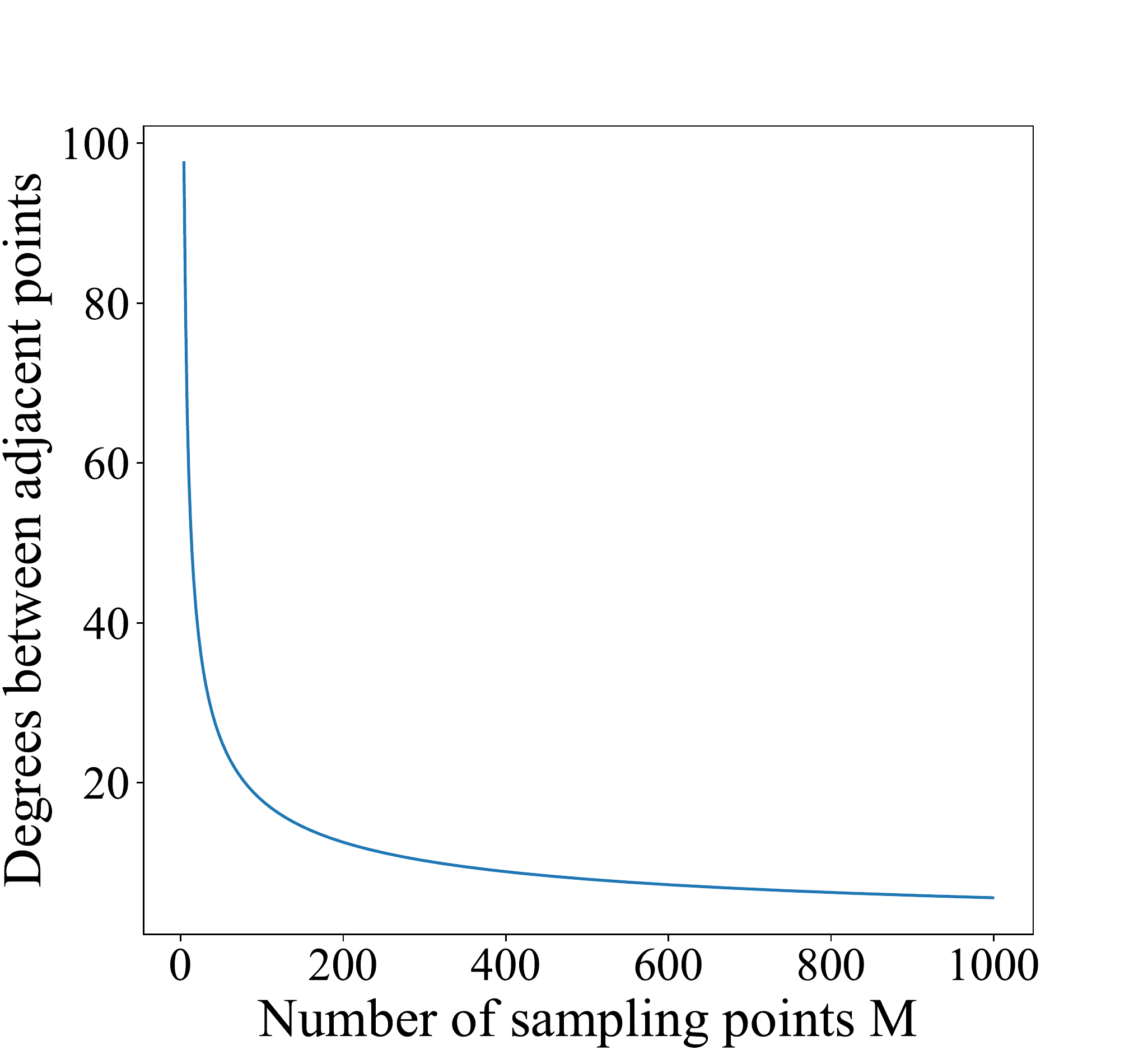}\\
    (a) \qquad \qquad \qquad \qquad \qquad \qquad \qquad (b) \qquad \qquad
\caption{\textbf{(a)} Visualization of the angle between two adjacent points. \textbf{(b)} Relationship between the number of sampling points $M$ and the angle between two adjacent points.}\label{fig:M_Angle}
\end{center}
\end{figure}

\section{Experiments}

\subsection{Datasets and Metrics}

We conduct an extensive set of experiments to evaluate our approach on three benchmarks: 300W-LP~\cite{zhu2016face}, AFLW2000~\cite{zhu2015high} and BIWI~\cite{fanelli2013random}. \textbf{300W-LP} is a synthesized dataset which contains 122,450 images with large varieties in facial poses and identities. Image samples in 300W-LP are synthesized from 300W dataset \cite{sagonas2013300} which includes around 4,000 images. \textbf{AFLW2000} contains the first 2,000 images of the popular AFLW~\cite{koestinger11a} dataset with diverse facial poses in the wild. The dataset is commonly used as the test set to evaluate model performances. \textbf{BIWI} is collected in an indoor environment with an RGB-D camera. It provides accurate ground truth labels. This dataset is also widely used for depth-based facial pose estimation. Since bounding boxes of human heads are not provided in BIWI, we use MTCNN \cite{zhang2016joint} to detect and crop the face areas.\\
\indent To ensure a fair comparison with different methods, we follow the same experiment scenarios applied in \cite{hsu2018quatnet,ruiz2018fine,yang2019fsa} and discard the test samples with Euler angles beyond the range of $[-99^\circ, 99^\circ]$. \textbf{Scenario 1:} We train our network on 300W-LP and test on both AFLW2000 and BIWI datasets. \textbf{Scenario 2:} We perform the 3-fold cross validation on the BIWI dataset. We randomly split the BIWI dataset into $3$ groups. Each group contains $8$ videos and the videos of the same person appear only in one group. We use mean absolute error of Euler angles (MAE) as our metric.

\begin{table*}[t]
\begin{center}
\caption{MAE and MAEV results of \textbf{different representations} of rotation under scenario 1 and 2. All use the ResNet-18 as backbone. We highlight the \textbf{\textcolor{blue}{best}} results.}
\begin{tabular}{c|c|c|ccc|c|ccc|c}
\toprule
\multirow{2}{*}{\textbf{Train}} & \multirow{2}{*}{\textbf{Test}} & \multirow{2}{*}{\textbf{Representation}} & \multicolumn{4}{c|}{\textbf{Euler Angle Errors}} & \multicolumn{4}{c}{\textbf{Vector Errors}} \\
\cline{4-11}
\multicolumn{1}{c|}{} &\multicolumn{1}{c|}{} & \multicolumn{1}{c|}{} & \textbf{Pitch} & \textbf{Yaw} & \textbf{Roll} & \textbf{MAE} & \textbf{Left} & \textbf{down} & \textbf{front} & \textbf{MAEV} 
\\ \hline
\multirow{8}{*}{\rotatebox[origin=c]{90}{300W-LP}} & \multirow{4}{*}{\rotatebox[origin=c]{90}{\parbox[c]{1cm}{\centering AFLW 2000}}} & Euler angles & $6.36$ & $4.64$ & $4.84$ & $5.28$ & $6.71$ & $5.97$ & $7.62$ & $6.76$ \\
\multicolumn{1}{c|}{} &\multicolumn{1}{c|}{} & Lie algebra & $5.62$ & $3.92$ & $4.04$ & $4.52$ & $5.84$ & $5.13$ & $6.52$ & $5.83$ \\
\multicolumn{1}{c|}{} &\multicolumn{1}{c|}{} & Quaternion & $5.77$ & $4.01$ & $4.20$ & $4.66$ & $5.63$ & $5.62$ & $6.57$ & $5.94$ \\
\multicolumn{1}{c|}{} &\multicolumn{1}{c|}{} & Rotation matrix & $\mathbf{\textcolor{blue}{5.46}}$ & $\mathbf{\textcolor{blue}{3.71}}$ & $\mathbf{\textcolor{blue}{3.77}}$ & $\mathbf{\textcolor{blue}{4.31}}$ & $\mathbf{\textcolor{blue}{5.52}}$ & $\mathbf{\textcolor{blue}{4.97}}$ & $\mathbf{\textcolor{blue}{5.92}}$ & $\mathbf{\textcolor{blue}{5.47}}$\\ 
\cline{2-11} 
& \multirow{4}{*}{\rotatebox[origin=c]{90}{\parbox[c]{1cm}{\centering BIWI (all)}}} & Euler angles & $6.43$ & $4.22$ & $4.08$ & $4.91$ & $6.08$ & $5.72$ & $6.13$ & $5.98$\\
& & Lie algebra & $5.87$ & $\mathbf{\textcolor{blue}{3.39}}$ & $3.73$ & $4.33$ & $5.82$ & $5.66$ & $5.42$ & $5.63$\\
& & Quaternion & $6.11$ & $3.54$ & $\mathbf{\textcolor{blue}{3.61}}$ & $4.42$ & $5.79$ & $5.88$ & $5.61$ & $5.76$ \\
& & Rotation matrix & $\mathbf{\textcolor{blue}{5.43}}$ & $3.52$ & $3.63$ & $\mathbf{\textcolor{blue}{4.19}}$ & $\mathbf{\textcolor{blue}{5.74 }}$ & $\mathbf{\textcolor{blue}{5.10}}$ & $\mathbf{\textcolor{blue}{5.12}}$ & $\mathbf{\textcolor{blue}{5.32}}$ \\ 
 \midrule
\multirow{4}{*}{\rotatebox[origin=c]{90}{\parbox[c]{1cm}{\centering BIWI (70\%)}}} & \multirow{4}{*}{\rotatebox[origin=c]{90}{\parbox[c]{1cm}{\centering BIWI (30\%)}}} & Euler angles & $4.07$ & $3.76$ & $3.73$ & $3.85$ & $4.52$ & $4.89$ & $4.57$ & $4.66$\\
 & & Lie algebra & $3.46$ & $3.21$ & $3.11$ & $3.26$ & $4.31$ & $4.22$ & $4.18$ & $4.24$ \\
 & & Quaternion & $3.52$ & $3.35$ & $3.24$ & $3.37$ & $4.51$ & $4.32$ & $4.20$ & $4.34$  \\
 & & Rotation matrix & $\mathbf{\textcolor{blue}{3.08}}$ & $\mathbf{\textcolor{blue}{3.16}}$ & $\mathbf{\textcolor{blue}{3.01}}$ & $\mathbf{\textcolor{blue}{3.08}}$ & $\mathbf{\textcolor{blue}{4.12}}$ & $\mathbf{\textcolor{blue}{4.16}}$ & $\mathbf{\textcolor{blue}{4.02}}$ & $\mathbf{\textcolor{blue}{4.10}}$ \\
\bottomrule
\end{tabular}
\label{tab:representation_comparison}
\end{center}
\end{table*}

\subsection{Implementation Detail}
There are two hyper-parameters in our approach. One is the coefficient $\alpha$ for the regression loss term $\mathcal{L}_{reg}$. Another one is the number of sampled points $M$. Fig.~\ref{fig:M_Angle} shows the relationship between number of sampled points $M$ and angle between adjacent points. We set $\alpha=0.2$ and $M=600$ in our experiments. 
\indent We implement our proposed approach based on PyTorch and adopt ResNet-18~\cite{he2016deep} as the backbone. In training, we adopt Adam optimizer with the initial learning rate of 0.0001. The total training epoch is set to be 50 with the decay rate of 0.95 for every epoch. Batch size is set to be 64 and every image is resized to $224 \times 224$. All the experiments are conducted on a RTX 2080 Ti GPU. 
We augment training images with random crop, noise and random zoom with scale from $0.8$ to $1.2$.

\begin{table*}[t]
\begin{center}
\caption{Comparison with state-of-the-art methods on the AFLW2000 and BIWI datasets. All methods are trained on 300W-LP. We highlight the \textcolor{blue}{\textbf{best}} results and \textbf{our} results.}
\begin{tabular}{c|c|cccc|cccc}
\toprule
\multicolumn{1}{c|}{\multirow{2}{*}{\textbf{Method}}} & \multicolumn{1}{c|}{\multirow{2}{*}{\textbf{Backbone}}} & \multicolumn{4}{c|}{\textbf{AFLW2000}} & \multicolumn{4}{c}{\textbf{BIWI (full)}}\\ 
\cline{3-10}
\multicolumn{1}{c|}{} & \multicolumn{1}{c|}{} & \textbf{Pitch} & \textbf{Yaw} & \textbf{Roll} & \textbf{MAE} & \textbf{Pitch} & \textbf{Yaw} & \textbf{Roll} & \textbf{MAE}\\\cline{2-5} 
\hline
  3DDFA\cite{zhu2016face} & Two-stream& $27.09$ & $4.71$ & $28.43$ & $20.08$ & $41.90$ & $5.50$ & $13.22$ & $20.21$ \\
  Dlib\cite{kazemi2014one}& -  & $11.25$ & $8.49$ & $22.83$ & $14.19$ & $13.00$ & $11.86$ & $19.56$ & $14.81$ \\
  HPE\cite{huang2020improving}& ResNet-50  & $6.18$ & $4.87$ & $4.80$ & $5.28$ & $5.18$ & $4.57$ & $3.12$ & $4.29$ \\
  Hopenet\cite{ruiz2018fine}& ResNet-50 & $7.12$ & $5.31$ & $6.13$ & $6.19$ & $5.89$ & $6.01$ & $3.72$ & $5.20$ \\
  Quatnet\cite{hsu2018quatnet}& GoogLeNet & $5.62$ & $3.97$ & $3.92$ & $4.50$ & $5.49$ & $4.01$ & $2.94$ & $4.15$ \\
  Liu $et$ $al$.~\cite{liu2019facial} & ResNet-50 & $5.06$ & $\textbf{\textcolor{blue}{3.03}}$ & $3.68$ & $3.93$ & $5.61$ & $4.12$ & $3.15$ & $4.29$\\
  FSA-Net\cite{yang2019fsa}& SSR-Net & $6.34$ & $4.96$ & $4.78$ & $5.36$ & $5.21$ & $4.56$ & $3.07$ & $4.28$ \\
  TriNet\cite{Cao_2021_WACV}& ResNet-50 & $5.77$ & $4.20$ & $4.04$ & $4.67$ & $4.75$ & $\mathbf{\textcolor{blue}{3.05}}$ & $4.11$ & $3.97$ \\
  MNN\cite{valle2020multi} & Encoder-Decoder & $\textbf{\textcolor{blue}{4.69}}$ & $3.34$ & $3.48$ & $3.83$ & $4.61$ & $3.98$ & $\mathbf{\textcolor{blue}{2.39}}$ & $3.66$ \\
  img2pose\cite{albiero2020img2pose}& ResNet-18 & $5.03$ & $3.43$ & $3.28$ & $3.91$ & $3.55$ & $4.57$ & $3.24$ & $3.79$ \\
\midrule
  \textbf{Ours} & ResNet-18 & $\mathbf{4.74}$ & $\mathbf{3.08}$ & $\mathbf{\textcolor{blue}{3.11}}$ & $\mathbf{\textcolor{blue}{3.64}}$ & $\mathbf{\textcolor{blue}{3.52}}$ & $\mathbf{4.21}$ & $\mathbf{3.10}$ & $\mathbf{\textcolor{blue}{3.61}}$ \\ 
\bottomrule
\end{tabular}\label{tab:sota}
\end{center}
\end{table*}

\subsection{Analysis of Rotation Representations}
\label{analysis_rot_rep}


Even though there are multiple ways to describe a rotation and the most commonly used ones include Euler angles, quaternion, Lie algebra and rotation matrices, it remains under-studied that which representation is the best option for the task of facial pose estimation. \cite{Cao_2021_WACV} briefly discussed Euler Angle and quaternions. However, they omitted Lie algebra and did not provide any experimental support. We implement a thorough comparison between the performances of different representations using the same backbone of ResNet-18 (see Table~\ref{tab:representation_comparison}). Since MAE is not an accurate measure for profile faces, we also adopt mean absolute error of vectors (MAEV) to make a comprehensive comparison. Experiments show that rotation matrices achieve the best result among all representations under both scenarios. \\
\indent The experimental results accord with the continuity properties of each representation. As shown by the work~\cite{xiang2021eliminating,zhou2019continuity}, it needs at least $5$ dimensions to describe the rotation continuously, otherwise it incurs discontinuity issue similar to Euler angles. Both Euler angle and Lie algebra $\in \mathbb{R}^3$ and quaternion $\in \mathbb{R}^4$. Therefore, none of them can describe the rotation continuously. Here we include some cases when the phenomenons of discontinuity occur. For a unit quaternion $\boldsymbol{q} = w + x \boldsymbol{i} + y \boldsymbol{j} + z \boldsymbol{{k}}$, where $w^2 + x^2 + y^2 + z^2 = 1$. Then $(1, 0, 0, 0)$ and $(-1, 0, 0, 0)$ represents the same rotation. For Lie algebra $\mathfrak{so}(3)$ which is denoted by an anti-symmetric matrix $\boldsymbol \phi^\wedge$ where $\boldsymbol \phi = \theta \boldsymbol{a}$ and $\boldsymbol{a} \in \mathbb{R}^3$, $||\boldsymbol{a}||_2 = 1, \theta \in [-\pi, +\pi]$. For any $\boldsymbol{a}$, faces have similar appearances when $\theta$ approaches $\pi$ and $-\pi$. Therefore, the rotation matrix is the best representation in terms of the performance and continuity property.

\subsection{Comparison with State-of-the-Arts}
We compare the performance of our method with other state-of-the-art methods (see Table~\ref{tab:sota}) under scenario $1$. Since the training/test set division in scenario $2$ is arbitrary and thus is not adopted by methods \cite{albiero2020img2pose,valle2020multi}, we choose only scenario $1$ for comparison. The results of the compared methods are directly cited from their original papers. Liu $et$ $al.$~\cite{liu2019facial} is the first work that follows the distribution learning paradigm for wild pose estimation. Different from our work, they convert the Euler angles to $3$ Gaussian distributions with the same variance. Even though they use the ResNet-50 as the backbone which is deeper than our ResNet-18, our ASG-based distribution learning surpasses their performance.
\begin{figure}[t]
\begin{center}
    \includegraphics[width=0.95\textwidth]{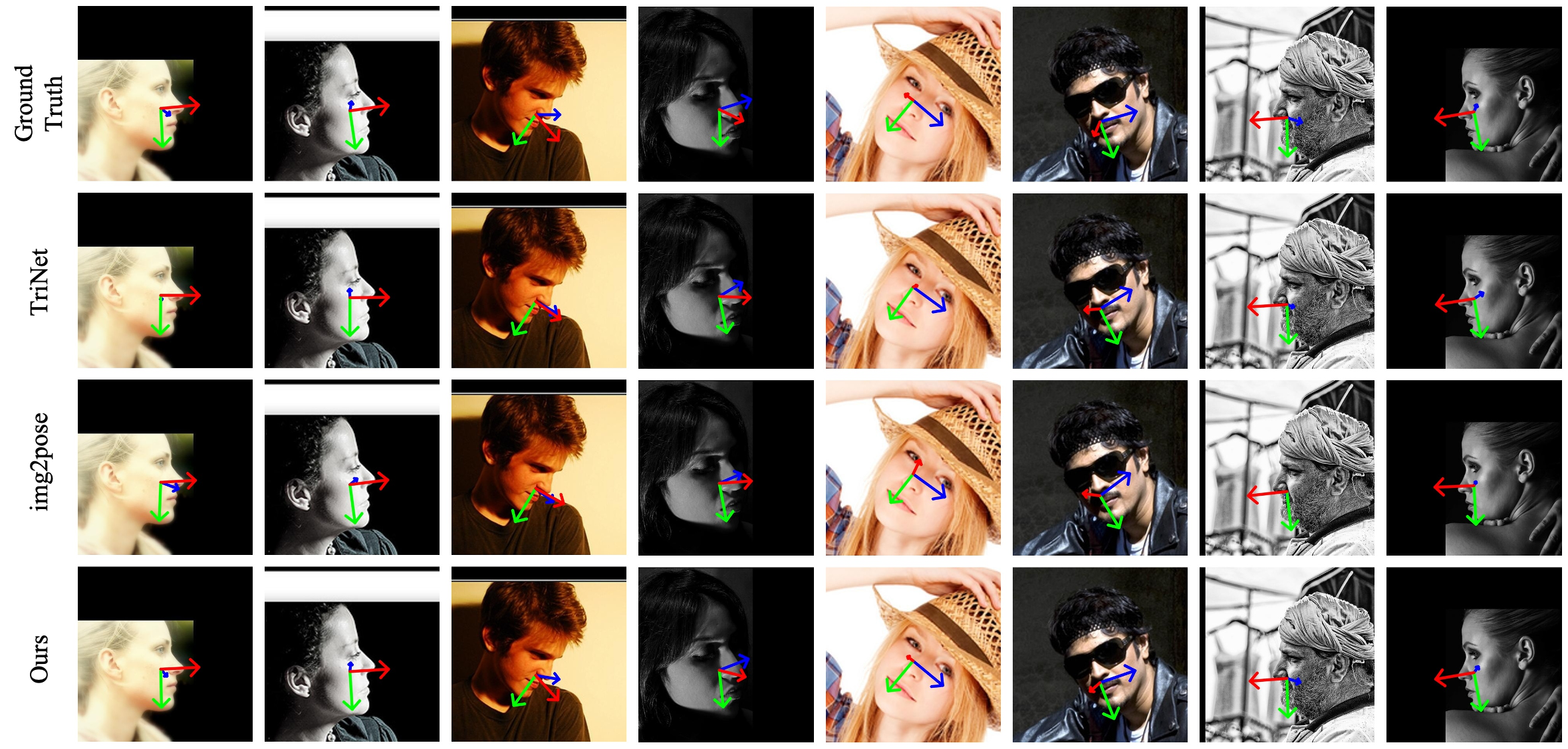}
\end{center}
\caption{\textbf{Qualitative comparison} of different methods. Trained on 300W-LP and tested on AFLW2000.}
\label{fig:qualitative_comparison}
\end{figure}
\begin{table}[h]
\begin{center}
\caption{Comparison between \textbf{direct regression} and \textbf{distribution learning}. Results are obtained on the AFLW2000 and BIWI benchmarks.}
\begin{tabular}{r|cc|c}
\toprule
\multicolumn{1}{r|}{Training set} & \multicolumn{2}{c|}{300W-LP} & \multicolumn{1}{c}{BIWI (70\%)} \\ 
\hline
\multicolumn{1}{r|}{Testing set}  & \multicolumn{1}{c}{ALFW2000} & \multicolumn{1}{c|}{BIWI (full)} & \multicolumn{1}{c}{BIWI (30\%)} \\ 
\hline
Direct regression & $4.31$ & $4.19$ & $3.08$ \\
SG learning & $3.79$ & $3.71$ & $2.93$\\
ASG learning & $\mathbf{3.64}$ & {$\mathbf{3.61}$} & $\mathbf{2.77}$ \\
\bottomrule
\end{tabular}
\label{tab:ablation_sg}
\end{center}
\end{table}
FSA-Net~\cite{yang2019fsa} and TriNet~\cite{Cao_2021_WACV} take advantage of the combination of attention module and capsule network and append them to the backbone network to improve the learning ability of the network. Even though both have more complex structures and more parameters than our network, their performance is inferior to ours.\\
\indent It is worth mentioning that some of the methods such as MNN~\cite{valle2020multi} and img2pose~\cite{albiero2020img2pose} also use face landmarks in a weakly supervised manner to help improve the performance of network. To highlight the effectiveness of our distribution learning strategy, we make the network learn the pose estimation without the landmark labels. Experiments show that even though less information is provided, our network still achieves better performance. The qualitative results are demonstrated in Fig.~\ref{fig:qualitative_comparison}. It can be seen that our approach makes more accurate predictions when faces are partially occluded. 
\begin{table}[t]
\begin{center}
\caption{Comparison between adaptive \textbf{parameters learning} and \textbf{fixed} ASG parameters. Results are obtained on the AFLW2000 and BIWI benchmarks.}
\begin{tabular}{r|cc|c}
\toprule
Training set & \multicolumn{2}{c|}{300W-LP} & BIWI (70\%) \\
\hline
Testing set & \multicolumn{1}{c}{AFLW2000} & \multicolumn{1}{c|}{BIWI (full)} & BIWI (30\%) \\ \hline
$\lambda = 1$, $\eta =1$ & $3.79$ & $3.84$ & $2.92$ \\
$\lambda = 5$, $\eta =5$ & $3.86$ & $3.97$ & $3.03$ \\
$\lambda = 1$, $\eta =5$ & $3.92$ & $4.03$ & $2.97$ \\
Adaptive Parameters & $\mathbf{3.64}$ & {$\mathbf{3.61}$} & $\mathbf{2.77}$\\ 
\bottomrule
\end{tabular}
\label{tab:ablation_fixed_parameters}
\end{center}
\end{table}
\begin{table}[t]
\begin{center}
\caption{Comparison of the effects of \textbf{different loss terms}. Results are obtained on the AFLW2000 and BIWI benchmarks.}
\begin{tabular}{r|cc|c}
\toprule
\multicolumn{1}{r|}{Training set} & \multicolumn{2}{c|}{300W-LP} & \multicolumn{1}{c}{BIWI (70\%)} \\ 
\hline
\multicolumn{1}{r|}{Testing set}  & \multicolumn{1}{c}{ALFW2000} & \multicolumn{1}{c|}{BIWI (full)} & \multicolumn{1}{c}{BIWI (30\%)} \\ \hline
$\mathcal{L}_{cls}$ & $3.67$ & $3.68$ & $2.81$ \\
$\mathcal{L}_{reg}$ & $4.31$ & $4.19$ & $3.08$ \\
$\mathcal{L}_{cls}$ + $\mathcal{L}_{reg}$ & $\mathbf{3.64}$ & {$\mathbf{3.61}$} & $\mathbf{2.77}$ \\ 
\bottomrule
\end{tabular}
\label{tab:ablation_losses}
\end{center}
\end{table}
\subsection{Ablation Study}
In this section, we investigate the effectiveness of our method by carrying out ablation experiments on the adaptive ASG label distribution learning and different loss components.
\begin{figure}[t]
  \begin{center}
    \includegraphics[clip,trim=0cm 2cm 0cm 7cm,width=0.75\textwidth]{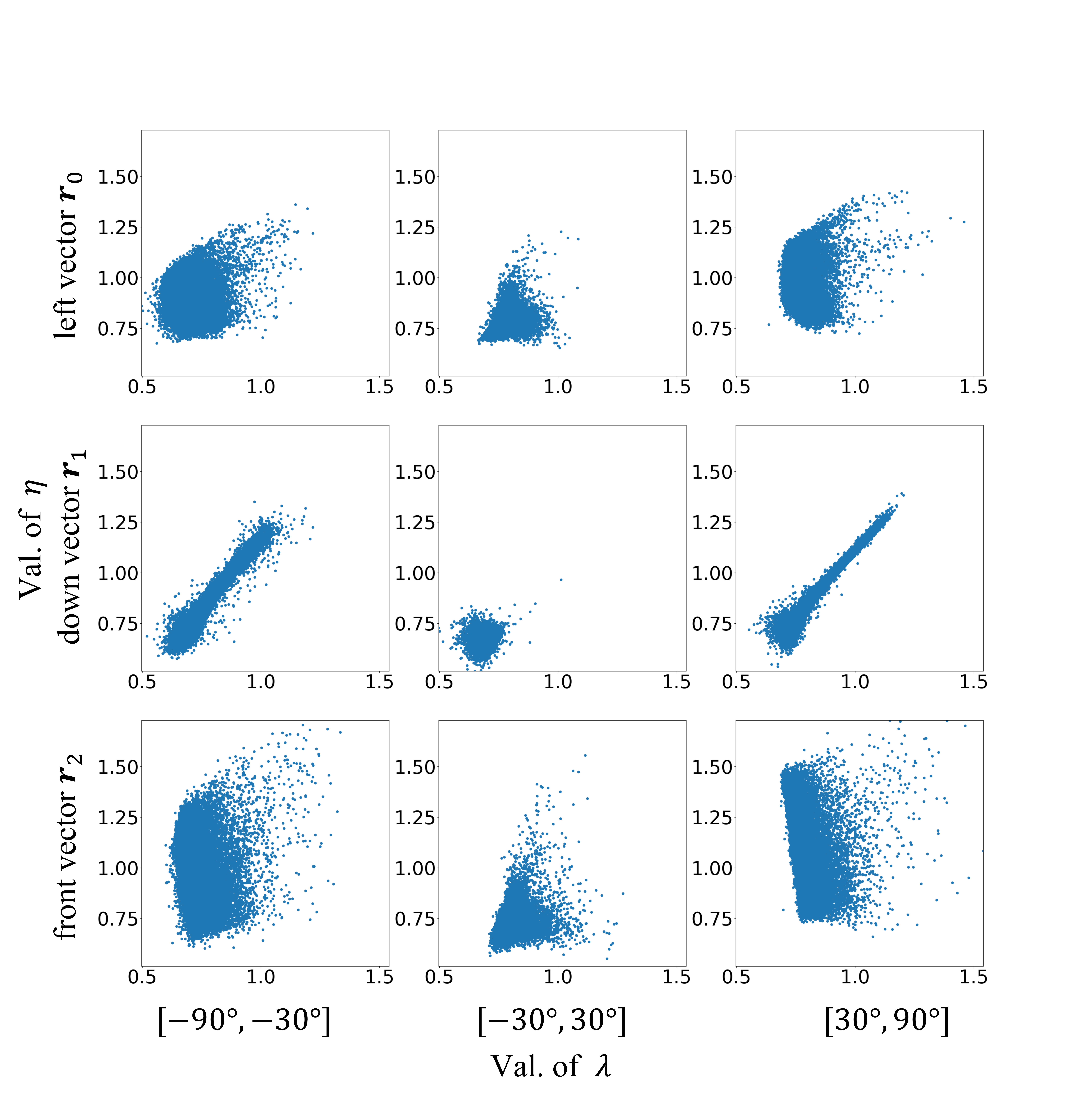}
  \end{center}
  \caption{Visualization of $\eta$ and $\lambda$ distribution for three pose vectors $\boldsymbol{r}_0, \boldsymbol{r}_1$ and $\boldsymbol{r}_2$ of all the samples in different ranges of Yaw. All trained on the 300W-LP dataset.}
  \label{fig:param_scatter}
\end{figure}

\noindent\textbf{Distribution learning $vs.$ regression.} We examine the advantages of the ASG distribution to isotropic SG distribution and use the direct regression of the rotation matrix as baseline. Results are shown in Table~\ref{tab:ablation_sg}. Our ASG distribution can effectively improve the performance compared with other two baseline methods. 

\noindent\textbf{Adaptive parameters $vs.$ fixed parameters.} We conduct experiments to compare the performance of methods with adaptive parameters and fixed parameters (see Table~\ref{tab:ablation_fixed_parameters}). While achieving superior performance over the fixed parameters, our adaptive parameter learning is  computationally efficient as it avoids the exhaustive search of parameters. All the ASG parameters $\lambda$ and $\eta$ learned by the samples in the 300W-LP dataset are demonstrated in Fig.~\ref{fig:param_scatter}. We divide the angle of yaw into three equal ranges. \textcolor{black}{It is worth noting that our learning behavior follows a clear pattern.} For instance, the parameter distributions in the first and third columns resemble each other. Because turning faces to left and right results in symmetric images, the parameters of ASG should be similar. This is reflected by the distribution of parameters.

\noindent\textbf{Loss functions.} We examine the effectiveness of each loss term (see Table~\ref{tab:ablation_losses}). Notice when only $\mathcal{L}_{reg}$ is applied, the network is supervised by the arbitrary distribution with expectation of the same as the ground truth. The above results  confirm that the classification term and regression term can work collaboratively to operate effective label distribution learning for the facial pose estimation.

\section{Conclusion}
In this paper, we introduce a novel ASG-based Label Distribution Learning method for estimating facial pose. This is the first attempt to include directional statistics in the estimation of pose. We anticipate that this work will illustrate potential future directions for the community to investigate.

\bibliographystyle{splncs04}
\bibliography{egbib}
\end{document}